\documentclass[journal,twoside,web]{ieeecolor}
\usepackage{generic}
\usepackage{cite}
\usepackage{amsmath,amssymb,amsfonts}
\usepackage{graphicx}
\usepackage{textcomp}
\usepackage{multirow}
\usepackage{amsmath}
\usepackage{stfloats}
\usepackage{color}

\usepackage{mathrsfs}

\usepackage{easyReview}
\usepackage[switch]{lineno}    
\usepackage[ruled]{algorithm2e}

\def\BibTeX{{\rm B\kern-.05em{\sc i\kern-.025em b}\kern-.08em
    T\kern-.1667em\lower.7ex\hbox{E}\kern-.125emX}}
\begin{document}
\title{Label-noise-tolerant medical image classification via self-attention and self-supervised learning}

\author{Hongyang Jiang, Mengdi Gao, Yan Hu, Qiushi Ren, Zhaoheng Xie, Jiang Liu  
\thanks{This work was supported in part by General Program of National Natural Science Foundation of China under Grant 82272086, in part by Shenzhen Science and Technology Program under Grant KQTD20180412181221912 and Grant JCYJ20200109140603831, \textit{(Hongyang Jiang and Mengdi Gao contributed equally to this work.)} \textit{(Corresponding author:  Zhaoheng Xie; Jiang Liu.)}}
\thanks{Hongyang Jiang, Yan Hu, and Jiang Liu are with the Department of Computer Science and Engineering, Southern University of Science and Technology, Shenzhen, China (e-mail: jianghy3@sustech.edu.cn; huy3@sustech.edu.cn; liuj@sustech.edu.cn).}
\thanks{Mengdi Gao and Qiushi Ren are with the Department of Biomedical Engineering, College of Future Technology, Peking University, Beijing 100871, China (e-mail: gaomengdi@pku.edu.cn; renqsh@coe.pku.edu.cn).}
\thanks{Zhaoheng Xie is with the Institute of Medical Technology, Peking University Health Science Center, Peking University, Beijing 100191, China (e-mail: xiezhaoheng@pku.edu.cn).}
}
\maketitle

\begin{abstract}
Deep neural networks (DNNs) have been widely applied in medical image classification and achieve remarkable classification performance. These achievements heavily depend on large-scale accurately annotated training data. However, label noise is inevitably introduced in the medical image annotation, as the labeling process heavily relies on the expertise and experience of annotators. Meanwhile, DNNs suffer from overfitting noisy labels, degrading the performance of models. Therefore, in this work, we innovatively devise noise-robust training approach to mitigate the adverse effects of noisy labels in medical image classification. Specifically, we incorporate contrastive learning and intra-group attention mixup strategies into the vanilla supervised learning. The contrastive learning for feature extractor helps to enhance visual representation of DNNs. The intra-group attention mixup module constructs groups and assigns self-attention weights for group-wise samples, and subsequently interpolates massive noisy-suppressed samples through weighted mixup operation. We conduct comparative experiments on both synthetic and real-world noisy medical datasets under various noise levels. Rigorous experiments validate that our noise-robust method with contrastive learning and attention mixup can effectively handle with label noise, and is superior to state-of-the-art methods. 
An ablation study also shows that both components contribute to boost model performance. The proposed method demonstrates its capability of curb label noise and has certain potential toward real-world clinic applications.
\end{abstract}


\section{Introduction}
\label{sec:introduction}

\IEEEPARstart{O}{ver} the past decade, artificial intelligence especially deep neural networks (DNNs) have been widely applied in medical image classification tasks. DNNs have exhibited impressive and often unprecedented performance, ranging from early screening to determining subcategories. The remarkable achievements have attributed to the increasing availability of large-scale medical datasets and powerful computational hardware. Normally, large quantity of medical data with reliable labels are favorable to train DNNs. However, it is time-consuming and laborious to accumulate and annotate vast amounts of medical data due to the data privacy and professional annotation requirements. Furthermore, label noise is inevitably introduced when labeling the medical images. Generally, contents of medical images tend to be of fine-granularity and their categories or grades are identified through the number (size and position) of lesions or pathological changes. Therefore, tasks for labeling medical images are challenging and annotators must possess relevant professional knowledge. It is a common phenomenon that noisy labels are in the presence of medical datasets \cite{karimi2020deep}. 
In addition, the frequently-used two current methods to collect medical data and labels would also introduce noisy labels with varying noise ratios. The first is to gather more training data with labels through crowdsourcing \cite{cheplygina2018crowd} or digging into the existing clinical reports \cite{wang2018tienet}. The other is to utilize semi-supervised learning framework \cite{wang2020seminll}, such as pseudo labels \cite{arazo2020pseudo}, to excavate potential information from unlabeled medical data. Notably, data with labels collected from crowdsourcing or pseudo labels suffer from heavy noise and thus these techniques have limited applicability in medical imaging \cite{kuznetsova2020open}. Relatively small datasets with noisy labels are common scenarios in the medical imaging applications \cite{dgani2018training}. 

It is universally acknowledged that DNNs possess powerful mapping capability and they can easily fit the entire training dataset even with any ratio of corrupted labels \cite{park2021provable}. The memorizing of noise data leads to poor generalization on the test dataset and the generalization performance degrades as the noise ratio increases. Consequently, noise-resistant training methods are highly desired when handling medical image classification tasks. For example, a promising solution to curb adverse effects of noisy labels is to train DNNs on those cherry-picking samples with small losses. Previous literature \cite{park2021provable} has demonstrated that DNNs tend to prioritize learning simple patterns first and then memorize the remaining data, including noisy data. Based on this observation, Co-teaching \cite{han2018co}, the representative of co-training with dual-network, selected reliable training samples to cross-update the peer network simultaneously. It can discriminate probably clean samples and prevent classifiers overfitting the noisy ones. However, its defects of discarding unreliable samples, including some hard but correct samples which are significant to the overall performance. Moreover, Co-teaching is highly dependent of the noise ratio to determine the portion of selected instances. In fact, we cannot forecast the prior noise ratio in real-world noisy medical images classification scenarios. Although noise ratio can be inferred using cross-validation according to the previous research \cite{liu2015classification}, estimation error of noise ratio is inevitably introduced, which is eventually adverse to generalization performance of DNNs. Finally, the dual-network style methods double the quantity of parameters, raising the computing resources and training time considerably.

In this paper, to address these issues, we proposed an end-to-end noise-robust DNNs framework, which is specifically designed for medical image classification tasks. The framework allows for a full exploration of the training data including the noisy data and does not require any prior knowledge, such as the noise ratio. Moreover, the framework possesses a shared feature encoder with two exclusive heads, reducing the number of model parameters. The framework consists of two core components. Firstly, intra-class mixup module can estimate the confidence scores of all the images in the noisy subset and generate a clean representation from the noisy subset, suppressing the influence of mislabeled data. Secondly, a two-stage joint loss function is devised, which is integrated intrinsic similarity loss for self-supervised module and mixup supervised loss for noise-suppressed module.

The proposed anti-noise framework could take advantage of all the training data including the noisy data, and handle various type of simulation noises with varying rates. We conduced comparative experiments to validate the superiority of our methods on three synthetic medical datasets and one real-world natural dataset. Compared with the state-of-the-art methods, ours achieves the best performance at the most cases in which noise ratios range from 10\% to 40\%. In summary, the contributions of this paper are as follows:
\begin{enumerate}[]
	\item This work proposes a noise-robust DNNs framework for medical image classification based on intra-class mixup and self-supervised techniques. The framework is independent of backbone network and prior information (such as, noise ratio and clean validation dataset).
	\item We presented a mini-group batch sampling criteria for the attentive noisy feature mixup module, which effectively suppressed the influence of noisy samples during the training.
	\item Not only simulated noisy public multi-modality medical datasets, but also a real-world dataset validates the superiority of our proposed method.
	\item Two types of synthetic noise: instance-independent and instance-dependent label noise, are manually introduced into the noise-free medical datasets. Our method is proved to resistant to different noise levels.
	\item Comprehensive experimental results witness the state-of-the-art classification performance of our proposed method under different types of noise with different ratios.	
\end{enumerate}

\section{Related work}
\subsection{Learning from noisy labels}
Research literature on learning from noisy labels exhibits great diversity in the computer vision community, roughly spanning from noise-robust to noise-cleaning methods. 
First, noise-robust methods are assumed to be insensitive to label noise, which directly trains DNNs in the presence of noisy labeled data. 
Some works address the issue of noisy label by designing  specific DNNs architectures, including developing a dedicated architecture or adding a noise adaptation layer at the top of the SoftMax layer \cite{goldberger2016training}. The resulting architectures yield improved generalization through the modification of the DNNs output based on the estimated label transition probability. However, the dedicated architecture lacks flexibility and the noise adaptation layer hinders a model’s generalization to complex label noise.
Instead of adjusting network structure, Jenni et. al. \cite{jenni2018deep} added regularization item in the expected training loss. Although the explicit regularization can lead to performance gain if they are properly tuned, it introduces sensitive model-dependent hyperparameters or requires deeper architectures to compensate for the reduced capacity.

Second, noise-cleaning methods aim to identify and then remove or refurbish noisy labels, cherry-picking clean samples to update DNNs. To prevent accumulated error caused by incorrect selection, recent approaches usually leveraged multiple DNNs to cooperate with one another. More concretely, Co-teaching \cite{han2018co} maintained two identical DNNs with corresponding random initialization parameters simultaneously. Each DNN was cross-updated through small-loss samples selected by its peer DNN. In addition to clean samples, researches attempted to selectively refurbish the noisy labels to further exploit the noisy samples as many as possible. SELFIE \cite{song2019selfie} utilized both the clean samples with small-loss and refurbished samples to update the DNNs. It reduced the possibility of the false correction while exploiting the more training data. Noise-cleaning methods can eliminate the noise accumulation through discarding some unreliable samples according to elaborate selection criterion. Although these methods have achieved significant improvements, they only conduct the partial exploration of the training data, excluding obscure yet useful training samples (hard samples) as well. Besides, these types of methods suffer from defect of heavy dependence on prior knowledge. 

\subsection{Self-supervised learning}
Self-supervised learning (SSL) can be viewed as a branch of unsupervised learning since it 
attempts to learn general representation from data without any human-annotated labels. SSL benefits almost all types of downstream tasks through leveraging annotation-free pretext tasks. Contrastive learning with instance discrimination task is dominant among the existing empirical SSL methods. Contrastive learning aims to group similar samples pairs closer and diverse samples pairs away from each other. And the similarity of feature representations in each pair is measured by intrinsic similarity metric. Current contrastive learning methods generally depends on a combination of intrinsic similarity and a sequence of image transformations. MoCo \cite{he2020momentum} constructs a dynamic dictionary with a memory bank and proposes a momentum encoder to obtain large and consistent dictionaries of visual representations. SimCLR \cite{chen2020simple}, with end-to-end training framework, illustrates the significance of diversity for data augmentation and proved that a nonlinear projection head can further boost the feature representation. More recently, BYOL \cite{grill2020bootstrap} proposed a non-contrastive learning framework which enforces the perturbation consistency between different views, and prevented mode collapsing by leveraging both the target network and the slow-moving average online network. The SSL can build a low-dimensional feature embedding space without relying professional medical annotation information, in which the structural similarity of the original dataset can be well preserved. Hence, the SSL is considered as one of the best choices for feature extraction module, especially when there are noisy data points. In this paper, the SSL was introduced in the whole training process, which helped to maintain the relationship between clean and noisy medical images in low-dimensional representation. The full exploration of all images including the noisy portion, can boost feature representations of encoders, make DNNs robust to label noise.

\begin{figure*}[!t]
	\centering
	\includegraphics[width=1.69\columnwidth]{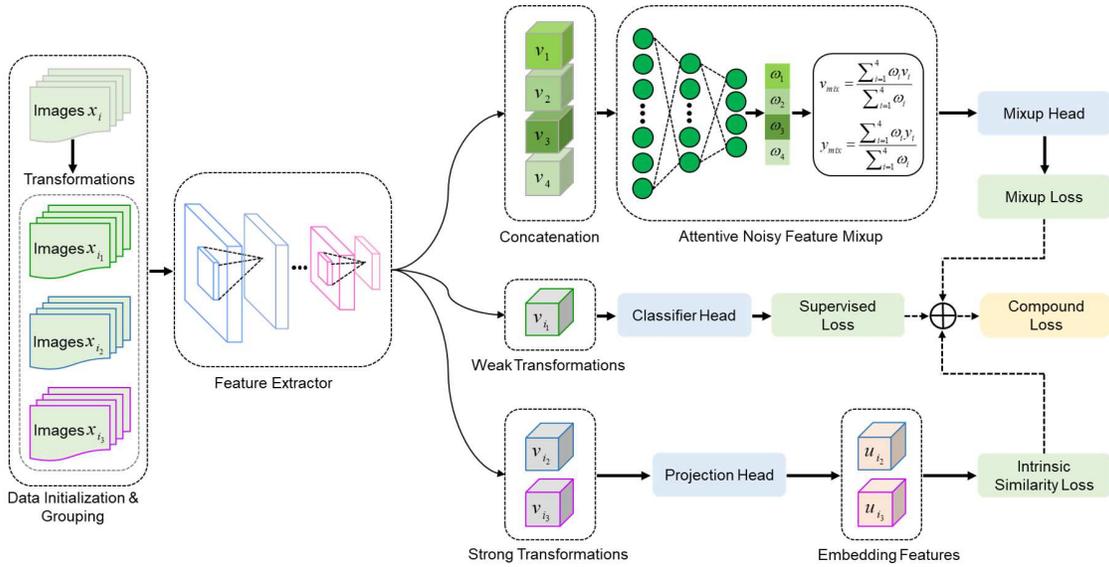}
	\vspace{-1.0em}
	\caption{The whole framework of our proposed noise-robust training method. In addition to conventional supervised loss, it originally integrates contrastive learning module for better visual representations and attentive noisy feature mixup module for reducing the influence of noisy label.}
	\label{fig:method}
	\vspace{-1.5em}
\end{figure*}

\section{MATERIAL AND METHODS}
\subsection{Problem Setting}
Medical image classification is a representative supervised learning task, seeking for the optimum mapping function $\mathcal{F}(\cdot;\theta)$ from the image space $X$ to label space $Y$. $X$ generally consists of a training dataset $\mathcal{D}_{train}={(\mathcal{X}_i,\mathcal{Y}_i)}_{i=1}^N$, where each pair $(\mathcal{X}_i,\mathcal{Y}_i)$ is independent and identically distributed and $N$ is the number of input-label data pairs. The optimal classifier is the one which minimizes the empirical risk.
\begin{equation}
	\mathcal{F}(\cdot;\theta)=\mathop{argmin}\limits_{\theta} \sum_{(\mathcal{X}_i,\mathcal{Y}_i)\in \mathcal{D}_{train}} \mathcal{L}(\mathcal{Y}_i,\mathcal{F}(\mathcal{X}_i,\theta))
	\label{eq_1}
\end{equation}
where $\mathcal{L}$ is a certain loss function and $\mathcal{F}(\mathcal{X}_i,\theta)$ is the predicted label of input $\mathcal{X}_i$. However, noisy labels extensively exist in the actual medical image classification tasks. Specifically, we are provided with a noisy training dataset ${\widetilde{\mathcal{D}}}_{train}={(\mathcal{X}_i,\tilde{\mathcal{Y}_i})}_{i=1}^N$ where $\tilde{\mathcal{Y}_i}$ is a noisy label (i.e., possibly incorrect). In the real situation, the ground truth of label $\mathcal{Y}_i$ is unknowable due to various annotation limitations, misdiagnosis, or disagreements. In those simulated noisy datasets, the prior ground truth labels of noisy data are still pretended unknown. Hence, the risk minimization process is unreliable, as noise labels change the original distribution of training dataset. In this situation, we aim to design a noise-tolerant training framework, which optimizes the DNNs with ${\widetilde{\mathcal{D}}}_{train}$ but achieves comparable performance based on $\mathcal{D}_{train}$. 

\subsection{Label-noise Representation Learning Framework}
To alleviate the error propagation coming from the training instances with incorrect labels, we originally integrate contrastive learning for better visual representations and attentive noisy feature mixup for reducing the influence of noise. Contrastive learning proceeds without the supervision of image labels. Consequently, it can fully utilize all the training samples and is immune from incorrect image labels. Meanwhile, intra-group attentive noisy feature mixup strategy is designed to generate `clean' features combination for the following classification through dynamically down-weighting the noisy samples in small groups. The construction of group is described in the following Section C.  
Compared to co-training-like methods with independent dual-network, our proposed method has a shared feature encoder with three exclusive heads, including classifier head ($\mathcal{C}(\cdot)$), projection head ($\mathcal{H}(\cdot)$), and mixup head ($\mathcal{M}(\cdot)$).
The noise-robust training framework is illustrated in \textcolor{blue}{Fig. \ref{fig:method}}.
Notably, our approach can be implemented conveniently without the need for such prior knowledge as noise rates, data distributions, and additional clean samples. The technical details are described in the following sections.

\subsection{Dataset Initialization and Grouping}
The whole dataset can restrain individual noise data points through internal similarity, that is the majority rule \cite{peng2020suppressing}. In this point, we propose a Mini-Group Batch Sampling (MGBS) criteria for attentive noisy feature mixup module. More specifically, we first divide the samples in each category into multiple mini-groups and each mini-group shares the identical given label. 
Then, we can obtain a mini-batch samples $\mathcal{X}_B={\{\mathcal{X}_{G_1}, \mathcal{X}_{G_2}, \cdots, \mathcal{X}_{G_K}\}}$ for each iteration, in which the size of $\mathcal{X}_{G_i}$ is $M$ and $i\in [1,K]$, and the actual batch size can be accumulated as $N_b=K\times M$. When $M=1$, MGBS is degenerated into the conventional batch sampling method. According to the majority rule \cite{peng2020suppressing}, the anti-noise performance of each mini-group is enhanced when the $M$ increases. However, a larger $M$ may damage the characteristics of independent data points. We will thoroughly analyze the size of $M$ in the discussion section.
 
Moreover, each sample $\mathcal{X}_i$ is transformed into three samples $(\mathcal{X}_{i_1}, \mathcal{X}_{i_2}, \mathcal{X}_{i_3})$ through different combinations of image augmentations. Specifically, $\mathcal{X}_{i_1}$ is the weak transformed image through the basic affine transformation including rotation, vertical and horizontal flip, with probability of 50\%. Meanwhile, $\mathcal{X}_{i_2}$ and $\mathcal{X}_{i_3}$ are the strong transformation versions based on both the weak transformation and some pixel-value transformations, supplementing gaussian filtering, color and grayscale jitter with 50\% probability. Notably, the color jitter is only used for three-channel medical images, and the grayscale jitter is for single channel medical images. To avoide meaningful medical content loss, we do not perform random crop or erasing operations. Here, $\mathcal{X}_{i_1}$ was fed into feature extractor, obtaining $\mathcal{V}_{i_1}$ and directly contributing to calculating supervised learning loss. $\mathcal{X}_{i_2}$ and $\mathcal{X}_{i_3}$ were fed into feature extractor, obtaining $\mathcal{V}_{i_2}$ and $\mathcal{V}_{i_3}$ and preparing for contrastive learning loss.

\subsection{Contrastive Learning for Feature Extractor}
Contrastive learning is a simplified self-supervised learning method without requiring specialized architectures and extra label memory \cite{jing2020self}. In the context of contrastive learning, we confirm a consensus that projections from the same image with different views and transformations should share the same label. We employ a projection function $\mathcal{H}(\cdot)$ to map $\mathcal{V}_{i_2}$ and $\mathcal{V}_{i_3}$ in high dimension space to $\mathcal{U}_{i_2}$ and $\mathcal{U}_{i_3}$ in low dimension space, respectively. $\mathcal{H}(\cdot)$ is a $\mathcal{Z}$-layer neural network and we investigate the one-layer and two-layer neural network in our study. The intrinsic similarity between $\mathcal{U}_{i_2}$ and $\mathcal{U}_{i_3}$ can be measured by a contrastive loss function as presented in the previous work \cite{oord2018representation}.

\begin{small}
\begin{equation}
    L(\mathcal{U}_{i_2},\mathcal{U}_{i_3})=-log\dfrac{exp(\mathcal{D}(\mathcal{U}_{i_2},\mathcal{U}_{i_3})/\tau)}{\sum_{j=1,i \neq j}^{N}\sum_{{t_i,t_j}\in\{2,3\}}exp(\mathcal{D}(\mathcal{U}_{i_{t_i}},\mathcal{U}_{j_{t_j}})/\tau)},
	\label{eq_2}
\end{equation}
\end{small}

where, $\mathcal{D}(\mathcal{U}_{i_2},\mathcal{U}_{i_3})$ refers to the pairwise similarity between $\mathcal{U}_{i_2}$ and $\mathcal{U}_{i_3}$, which is defined in (\textcolor{blue}{eq. \ref{eq_3}}). $\tau$ is a temperature hyper-parameter that is set to 0.5 in our study. The contrastive learning can be guided by intrinsic similarity loss $\mathcal{L}_c$ that is defined in (\textcolor{blue}{eq. \ref{eq_4}}).
\begin{equation}
	\mathcal{D}(\mathcal{U}_{i_2},\mathcal{U}_{i_3})=\frac{\mathcal{U}_{i_2}^T\mathcal{U}_{i_3}}{\Vert\mathcal{U}_{i_2}\Vert \Vert\mathcal{U}_{i_3}\Vert}
	\label{eq_3}
\end{equation}
\begin{equation}
	\mathcal{L}_c=\sum_{i=1}^{N}(L(\mathcal{U}_{i_2},\mathcal{U}_{i_3})+L(\mathcal{U}_{i_3},\mathcal{U}_{i_2}) )
	\label{eq_4}
\end{equation}

\subsection{Intra-group Attentive Noisy Feature Mixup}
To further depress the noisy-labeled samples in a mini-group, we proposed an Intra-group Attentive noisy Feature Mixup (IAFM) module to perform weighted fusion of samples based on MGBS criteria. To be specific, IAFM initially estimates corresponding confidence score for each sample in mini-group $X_{G_i}(i=[1,K])$, and then a relatively `clean' feature representation can be generated by summarizing intra-group features $\mathcal{V}_{i}$ with their corresponding learnable attention weights ($w_i$). The resulting feature representation from IAFM can be defined as follows,
\begin{equation}
	\mathcal{V}_{mix}=\dfrac{\sum_{i=1}^{K}w_i\mathcal{V}_{i}}{\sum_{i=1}^{K}w_i},
	\label{eq_5}
\end{equation}
where $w_i$ and $\mathcal{V}_{i}$ are confidence score and feature embedding for the $i^{th}$ sample in a mini-group, respectively. The IAFM module can dynamically allocate the contribution from each sample in a group during each training iteration, which encourages clean and noisy samples interact with each other. Hence, with the help of IAFM, less-deviated feature representations can be learned for the down-stream decision-making task. The ground-truth of mixup representation $\mathcal{V}_{mix}$ can be calculated as:
\begin{equation}
	\mathcal{Y}_{mix}=\dfrac{\sum_{i=1}^{K}w_i\mathcal{Y}_{i}}{\sum_{i=1}^{K}w_i}
	\label{eq_6}
\end{equation}
In our study, samples in a mini-group are from the same category, so that the $\mathcal{V}_{mix}$ inherits the corresponding label, i.e., $\mathcal{Y}_{mix}=\mathcal{Y}_{1}=\mathcal{Y}_{2}=\cdots=\mathcal{Y}_{K}$. Then, the $\mathcal{V}_{mix}$ can be fed into the last classifier and the group-pairs $(\mathcal{V}_{mix},\mathcal{Y}_{mix})$ are used for optimizing the classifier. Although the $\mathcal{V}_{mix}$ weakens the influence of noisy samples, it smooths the independent characteristics of real samples. Hence, we also retain the supervised learning of independent sample pairs $(\mathcal{X}_i, \mathcal{Y}_i)$. Finally, the classifier is updated through a weighted decision loss function:
\begin{equation}
	\mathcal{L}_{d}=\frac{1}{\sigma_1}\mathcal{L}_{m}+\frac{1}{\sigma_2}\mathcal{L}_{s}+log(\sigma_1\sigma_2),
	\label{eq_7}
\end{equation}
where $\mathcal{L}_{m}$ and $\mathcal{L}_{s}$ are vanilla cross-entropy loss function for $(\mathcal{V}_{mix},\mathcal{Y}_{mix})$ and $(\mathcal{V}_{i},\mathcal{Y}_{i})$ respectively. Further, $\sigma_{1}$ and $\sigma_{2}$  are learnable hyper-parameters which are updated in each training iteration. Their initial values are set to 1.0.

\subsection{Overall Loss Functions}
Our proposed approach utilizes a compound loss function with three
different terms. They are essentially contrastive learning loss $\mathcal{L}_c$, mixup loss $\mathcal{L}_{m}$, and supervised loss $\mathcal{L}_{s}$. In the first stage that is also the warm-up period, the contrastive learning loss function is solely exerted to guide the updating procedure of the feature extractor. Namely, the loss function $\mathcal{L}_{stage_1}$ is expressed in \textcolor{blue}{eq. \ref{eq_8}}.
At this stage, the intrinsic similarity between each data point in the whole dataset is excavated, which enhances the stability and noise resistance of feature extractor. 
In the second stage, in additon to contrastive learning loss, the IAFM is further exerted to reinforce the vanilla supervised learning, which is beneficial to obtain a noise-tolerant framework. 
More concretely, the weighted decision loss function plays a leading role, and the $\mathcal{L}_{c}$ can be viewed as a regularization over the whole training. As a result, the loss function is summarized as $\mathcal{L}_{stage_2}$ in \textcolor{blue}{eq. \ref{eq_9}}. Here, $\lambda$ is the regularization coefficient that is set to 0.1.
\begin{equation}
	\mathcal{L}_{stage_1} = \mathcal{L}_{c} 
	\label{eq_8}
\end{equation}
\begin{equation}
	\mathcal{L}_{stage_2} = \mathcal{L}_{d} + \lambda\mathcal{L}_{c} 
	\label{eq_9}
\end{equation}

\subsection{Dataset Accumulation and Transformation}
\textbf{Dataset description.} In this study, we altogether validated our method on three public avaliable medical datasets, Retina OCT \cite{kermany2018identifying}, Blood Cell \cite{acevedo2020dataset} and Colon Pathology \cite{kather2019predicting}. We reformatted the Retina OCT dataset as a two-class image classification task, covering normal and abnormal categories. The abnormal consists of choroidal neovascularization, diabetic macular edema, and drusen cases and the quantities of them are in balance. Due to the simplicity of the binary classification task, we randomly sampled 600 images as training dataset and 1500 images as test dataset for performance evaluation. The Blood Cell dataset was re-constructed on a prior database of individual cells which are originally divided into eight classes. We omitted one category with several sub-types and selected randomly 600 images for each category from the remaining seven types. Among this reformatted dataset, one third was for training and two thirds for testing. The Colon Pathology dataset with more categories were also employed in our study. We randomly sampled 1000 and 250 images of each category from the whole nine types of tissues as training and test dataset, respectively. In addition, the real-world ANIMAL10N dataset \cite{song2019selfie} was utilized to verify the generalization of our method. The date volume of training and testing and the image size of each dataset are demonstrated in \textcolor{blue}{Table \ref{tab: dataset}}.

\begin{table}[t!]
	\caption{Summary of dataset used in the experiment.}
	\begin{center}
		\label{tab: dataset}
		\vspace{-1.1em}
		\resizebox{0.90 \columnwidth}{9mm}{
			\begin{tabular}{ccccc} 
				\hline
				\hline
				Dataset & \# of training & \# of test & \# of class & size \\ 
				\hline				
				Retina OCT & 600 & 1500 & 2 & 224 $\times$ 224 \\
				Blood Cell & 1400 & 2800 & 7 & 112 $\times$ 112 \\
				Colon Pathology & 9000 & 2250 & 9 & 112 $\times$ 112 \\
				ANIMAL10N & 50000 & 5000 & 10 & 64 $\times$ 64 \\	
				\hline
				\hline
			\end{tabular}
		}
	\end{center}
	\vspace{-2.0em}
\end{table}

\textbf{Noise injection.} Since our datasets are with ground-truth labels, we need to corrupt the original labels manually to simulate noisy labels. Referring to previous study \cite{song2022learning}, we considered two types of synthetic noise: instance-independent and instance-dependent label noise. Inside, instance-independent label noise consists of symmetric noise and asymmetric noise. More concretely, symmetric noise refers that noisy labels are uniformly distributed among all categories with equal probability $P/\left|C\right|$. $P$ is noise rate and $P\in\left[0,1\right]$, and $C$ is the number of categories. Asymmetric noise refers that noisy labels are generated by flipping specific class to the similar class or the next class circularly with probability $P$. Hence, instance-independent label noise can be constructed by the label transition matrix which describes the probability of ground-truth label being flipped to the noisy label. 
Furthermore, instance-dependent label noise is investigated, which is a more realistic noise modeling. The corruption probability is assumed to be dependent on both the images features and class labels. First, we utilized training dataset to train DNNs, then we picked model with training accuracy nearly $1-P$ to get predictions of each training sample. Afterwards the predicted labels of training samples were considered as given labels. Consequently, the training labels were corrupted according to its preset noise rate.

In our work, we evaluated the robustness of our approach on varying noise rates from light noise to heavy noise. Referring to the real-world noise rate \cite{song2022learning} and previous work \cite{liu2021co}, five noise rates varying from 0\% to 40\% with step 10\% were simulated for all the types of noise model, to validate the robustness of our proposed method. Note that, noise rate is set as no more than 40\% because certain classes become theoretically indistinguishable on the condition that asymmetric noise is larger than 50\%.

\subsection{Comparative study methods}
We compared our method with a benchmark model (marked as Default) and the following state-of-the-art robust training methods (Label Smooth \cite{szegedy2016rethinking}, Co-teaching \cite{han2018co}, Co-teaching+ \cite{yu2019does}, JoCoR \cite{wei2020combating}, and Co-Correcting \cite{liu2021co}). We re-implemented all methods on the same network architecture (ResNet-18) by PyTorch. To ensure the fairness of the comparison, we adopted the same number of training epochs and the same optimization strategy. In addition, we re-adapted them with fine-tune hyperparameters on our newly fomated medical datasets. The comparison baselines are briefly introduced as follow:
\begin{enumerate}[]
\item Default refers to train DNNs directly on noisy datasets without any processing strategies of 
noisy labels, which is applied as a simple baseline.
\item Label Smooth can help DNNs prevent overfitting noisy labels through using soft labels 
rather than hard labels.
\item Co-teaching utilized dual-network to filter out small-loss samples respectively, then 
cross-updated the peer network through the selected small-loss samples.
\item Co-teaching+ inherited the advantages of dual-network structure and cross-updation pattern 
from Co-teaching. Besides, it replenished “Update by Disagreement” strategy which kept dual-network divergent.
\item JoCoR was to train dual-network simultaneously with small-loss instances based on a joint 
loss, including regular supervised loss and co-regularized loss.
\item Co-Correcting was a noise-tolerant framework, which improved accuracy and obtained more accurate labels through dual-network mutual learning, label probability estimation, and curriculum label correcting. 
\end{enumerate} 

\subsection{Implementation details} 
We utilized ResNet18 as the backbone network, which is standard unchanged architecture used in the paper \cite{he2016deep}, for all the experiments on Retinal OCT, Blood Cell, and Colon Pathology datasets. 
The training procedure utilized the Adam optimizer with an initial learning rate of 0.001 and a momentum of 0.9. The mini batch size was fixed as 8. For each trial, it took 30 epochs to conduct the first stage training and 70 epochs for the second stage training. Meanwhile, the learning rate began to decline by 0.1 times per 10 epochs from the second stage training.
All experiments were performed on an NVIDIA RTX3090 GPU with 24 GB memory. 

\section{EXPERIMENT AND RESULTS}
\subsection{Experiments on instance-independent label noise}
This section introduces the overall comparisons to related works, on the Retina OCT, Blood Cell, and Colon Pathology datasets. Note that, in support of reliable evaluation, we reported the average test accuracy of the last three epochs. As for other evaluation metrics, we chose the model closest to the average performance in the last three models, to conduct the comparative study. From default method in \textcolor{blue}{Table \ref{tab: accuracy_OCT}, \ref{tab: accuracy_HE}, and \ref{tab: accuracy_blood}}, it can be observed that DNNs are capable of memorizing noisy training dataset, leading to poor generalization on the test dataset. In addition, the generalization performance degrades as the noise ratio increases. Thereby, it is significant to mitigate label noise to boost the generalization performance of DNNs in medical image classification.

\textbf{Retina OCT dataset.} For binary classification on Retinal OCT dataset, asymmetric noise is degenerated into symmetric noise model. Therefore, we merely conducted experiments with symmetric noise and instance-dependent label noise. \textcolor{blue}{Table \ref{tab: accuracy_OCT}} presents the test accuracy of all comparison methods on Retinal OCT dataset. We can see that although all anti-noise comparative methods are capable of reducing the adverse effects of noisy labels, our method achieves top performance under various noise ratios. Even under severe noise, our method shows impressive performance which outperforms comparative methods by quite large margins. For example, ours has test accuracy of 85.18\% that is 29.15\% higher than 56.03\% from Default. More specifically, Label Smooth plays a limited role in curbing the adverse effects of noisy data and it tends to be complementary regularization strategy. In contrast sample-selection methods which discard all unclean samples, our method not only adapts self-supervised strategy to excavates valuable information from noisy data, but also adapts mixup strategy to mitigate effects of label noise. 
Another interesting phenomenon is that the accuracy of our method exceeds that of Default even on clean labels. Our proposed approach can not only deal with the noisy labels but also can boost the performance on noise-free data. 

\begin{table*}[t!]
	\caption{The test accuracy on Retina OCT dataset. Note that we report the average results of the last three epochs. 
	}
	\begin{center}
		\label{tab: accuracy_OCT}
		\vspace{-1.5em}
		\resizebox{1.6 \columnwidth}{11.5mm}{
			\begin{tabular}{cccccccc} 
				\hline
				\hline
				Noise Ratio & Default & Label Smooth & Co-teaching & Co-teaching$+$ & JoCoR & Co-Correcting & Ours \\ 
				\hline				
				0 & 95.56 & 95.63 & 95.49 & 94.88 & 94.76 & 93.27 & \textbf{96.75} \\
				0.1 & 85.94 & 86.32 & \textbf{94.33} & 92.41 & 90.90 & 89.42 & 94.02 \\
				0.2 & 80.26 & 80.81 & 89.74 & 88.32 & 89.46 & 89.77 & \textbf{91.24} \\
				0.3 & 74.31 & 74.80 & 84.07 & 85.38 & 85.34 & 79.90 & \textbf{89.56} \\
				0.4 & 56.03 & 58.89 & 51.27 & 64.46 & 72.99 & 55.48 & \textbf{85.18} \\								 			
				\hline
				\hline
			\end{tabular}
		}
	\end{center}
	\vspace{-1.5em}
\end{table*} 

\begin{table}[t!]
	\caption{Various metrics under the 30\% noise level on Retina OCT dataset.}
	\begin{center}
		\label{tab: metrics_OCT}
		\vspace{-1.5em}
		\resizebox{0.7 \columnwidth}{14mm}{
			\begin{tabular}{cccc} 
				\hline
				\hline
				Method & Precision & Recall & F1 Score \\ 
				\hline				
				Default & 76.89 & 69.20 & 72.84 \\
				Label Smooth & \textbf{94.51} & 52.80 & 67.75 \\
				Co-teaching & 85.81 & 81.47 & 83.58 \\
				Co-teaching$+$ & 91.56 & 79.60 & 85.16 \\
				JoCoR & 84.12 & 87.60 & 85.83 \\
				Co-Correcting & 89.37 & 68.40 & 77.49 \\
				Ours & 89.93 & \textbf{88.25} & \textbf{90.15} \\				 					 			
				\hline
				\hline
			\end{tabular}
		}
	\end{center}
	\vspace{-1.5em}
\end{table}

\begin{figure}[!t]
	\centering
	\includegraphics[width=0.9\columnwidth]{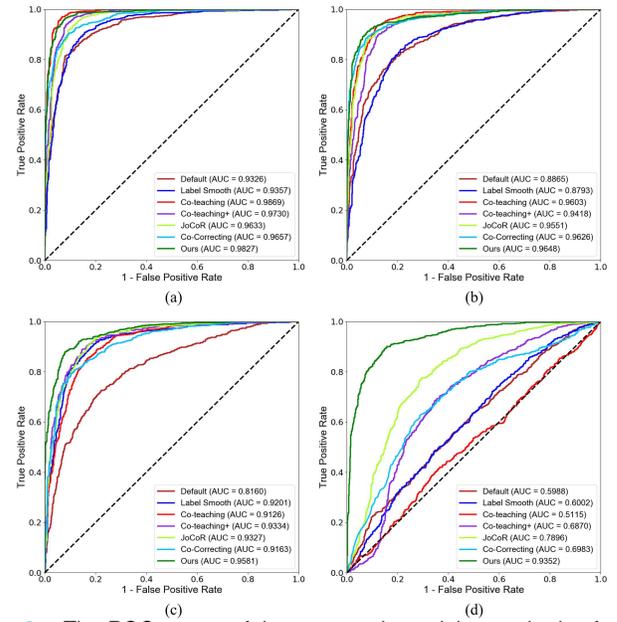}
	\vspace{-1.0em}
	\caption{The ROC curves of the comparative training methods of anti-noise on Retina OCT datasets under 10\% (a), 20\% (b), 30\% (c), and 40\% (d) noise settings, correspondingly.}
	\label{fig:ROC}
	\vspace{-1.0em}
\end{figure}

Furthermore, \textcolor{blue}{Table \ref{tab: metrics_OCT}} shows the precision, recall, and F1 score metrics of comparative study on Retina OCT dataset at the noise rate of 30\%. From this table, we again confirm that our method outperforms others. Our method achieves around a margin of 13.04\%, 19.05\%, and 17.31\% over the Default method on precision, recall, and F1 score metrics, respectively.
The last but not the least, \textcolor{blue}{Fig. \ref{fig:ROC}} shows the ROC curves of comparative methods on Retina OCT datasets under 30\% noise settings. Area Under Curve (AUC) values are marked in the legend correspondingly. As shown in \textcolor{blue}{Fig. \ref{fig:ROC}}, our proposed model possesses the best capability of recognizing normal and abnormal retina OCT images. AUC of our method achieves 95.81\%, which is 2.47\% higher than second-best Co-teaching+ method.

\textbf{Blood Cell dataset.} As for both blood cell dataset, we generated asymmetric noisy labels by flipping each class to the next class with noise rate, and conducted experiments with both instance-independent and instance-dependent label noise. The experimental results under different noise levels on Blood Cell dataset are listed in \textcolor{blue}{Table \ref{tab: accuracy_blood}}. We can observe that our method achieves the best result especially at heavy noise ratios. At the symmetric noise ratio of 40\%, our method far exceeds the Default method and surpasses the second-best JoCoR in the accuracy by 4.94 percentage points. At light noise ratio (10\%), our method is still competitive and obtains sub-optimal performance. Compared with binary classification task on Retina OCT dataset, the performance of anti-noise training methods when handling this multi-class classification task, degenerates more slowly as the noise ratio increases. This phenomenon is obvious especially for symmetric noise. We speculated that dominance of accurate category is scarcely influenced when noisy labels with varying noise ratios are diluted into multiple categories.

\begin{table*}[t!]
	\caption{The test accuracy on Blood Cell dataset. Note that we report the average results of the last three epochs.
	}
	\begin{center}
		\label{tab: accuracy_blood}
		\vspace{-1.5em}
		\resizebox{1.7 \columnwidth}{17.5mm}{
			\begin{tabular}{ccccccccc} 
				\hline
				\hline
				Noise Type & Noise Ratio & Default & Label Smooth & Co-teaching & Co-teaching$+$ & JoCoR & Co-Correcting & Ours \\ 
				\hline				
				Clean & 0 & 98.37 & 98.48 & 98.35 & 98.63 & \textbf{98.45} & 95.14 & 97.11 \\
				\hline
				\multirow{4}*{Symmetric noise} & 0.1 & 95.23 & 96.56 & 95.32 & \textbf{97.61} & 94.94 & 93.61 & 96.82  \\
				~ & 0.2 & 88.96 & 92.17 & 93.98 & 94.92 & 92.83 & 92.18 & \textbf{96.32} \\
				~ & 0.3 & 78.82 & 83.82 & 89.99 & 90.14 & 91.96 & 89.15 & \textbf{95.54}  \\
				~ & 0.4 & 68.48 & 74.86 & 89.46 & 82.63 & 90.27 & 84.7 & \textbf{95.21}  \\
				\hline
				\multirow{4}*{Asymmetric noise} & 0.1 & 93.04 & 95.63 & 92.2 & \textbf{96.17} & 92.32 & 92.8 & 96  \\
				~ & 0.2 & 84.57 & 90.27 & 90.1 & 91.66 & 91.76 & 89.5 & \textbf{94.32}  \\
				~ & 0.3 & 73.86 & 77.94 & 87.52 & 89.9 & 89.31 & 85.24 & \textbf{90.93}  \\
				~ & 0.4 & 63.75 & 63.9 & \textbf{85.2} & 77.02 & 80.4 & 81.61 & 80.75 \\ 								 			
				\hline
				\hline
			\end{tabular}
		}
	\end{center}
	\vspace{-1.7em}
\end{table*} 

\begin{table*}[t]
	\caption{The test accuracy on Colon Pathology dataset. Note that we report the average results of the last three epochs.
	}
	\begin{center}
		\label{tab: accuracy_HE}
		\vspace{-1.5em}
		\resizebox{1.7 \columnwidth}{17.5mm}{
			\begin{tabular}{ccccccccc} 
				\hline
				\hline
				Noise Type & Noise Ratio & Default & Label Smooth & Co-teaching & Co-teaching$+$ & JoCoR & Co-Correcting & Ours \\ 
				\hline				
				Clean & 0 & 91.1 & \textbf{91.5} & 89.99 & 89.78 & 90.7 & 84.8 & 91.06 \\
				\hline
				\multirow{4}*{Symmetric noise} & 0.1 & 79.75 & 80.62 & 85.45 & 88.02 & 80.43 & 80.66 & \textbf{89.41} \\
				~ & 0.2 & 71.41 & 79.8 & 85.54 & 86.82 & 77.52 & 78.73 & \textbf{89.9} \\
				~ & 0.3 & 60.81 & 67.79 & 83.79 & 79.13 & 73.98 & 77.46 & \textbf{88.52} \\
				~ & 0.4 & 57.05 & 57.86 & 72.9 & 75.22 & 71.95 & 76.37 & \textbf{87.28} \\
				\hline
				\multirow{4}*{Asymmetric noise} & 0.1 & 80.47 & 88.73 & 82.04 & 87.16 & 84.07 & 84.1 & \textbf{89.01} \\
				~ & 0.2 & 74.2 & 81.64 & 82.57 & 83.44 & 81.27 & 78.56 & \textbf{85.1} \\
				~ & 0.3 & 72.48 & 76.16 & 75.65 & 79.7 & 79.22 & 75.61 & \textbf{82.47}  \\
				~ & 0.4 & 60.28 & 62.62 & 73.03 & \textbf{78.67} & 77.65 & 66.19 & 74.42 \\ 								 			
				\hline
				\hline
			\end{tabular}
		}
	\end{center}
	\vspace{-2.2em}
\end{table*} 

\begin{figure*}[!t]
	\centering
	\includegraphics[width=1.95\columnwidth]{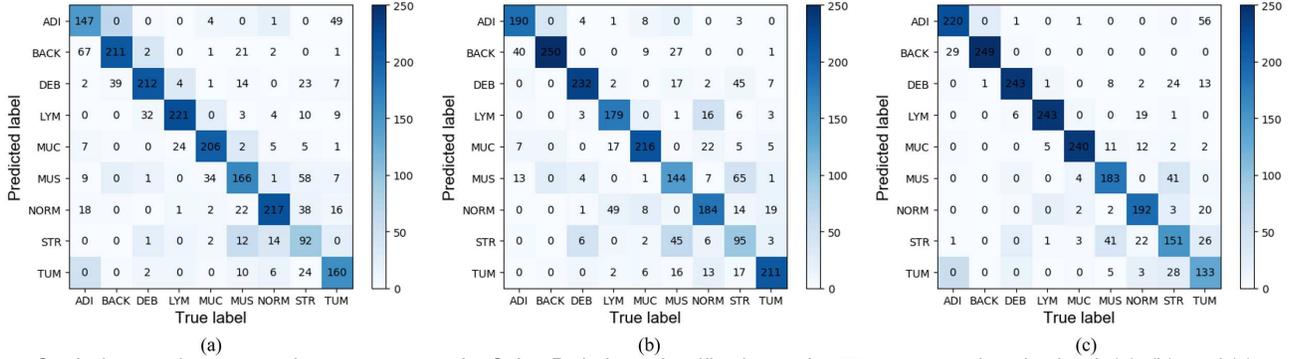}
	\vspace{-1.05em}
	\caption{Confusion matrices comparing test accuracy for Colon Pathology classification under 30\% asymmetric noise level. (a), (b), and (c) are for Default, co-correcting, and our method, respectively.}
	\label{fig:asymmetric30}
	\vspace{0.0em}
\end{figure*}

\textbf{Colon Pathology dataset.} Regarding colon pathology dataset which is a 10-class classification tasks, we generated asymmetric noisy labels in the aforementioned pattern, and still conducted experiments with three-type label noise. From \textcolor{blue}{Table \ref{tab: accuracy_HE}}, all comparative methods also excel in the classification tasks on Colon Pathology dataset, especially our method. Ours nearly achieves the optimal performance under different noise levels. Compared with the Default method, taking symmetric noise for example, ours achieves improvement on test accuracy of 9.66\%, 18.49\%, 27.71\%, and 30.23\% under the noise rate raising from 10\% to 40\%, respectively. Besides, the improvement of our method is more significant with the increase of noise ratio. At the heavy noise rate of 40\%, our method overpasses the sub-optimal Co-correct method with over 10\% gains. Label Smooth method slightly surpasses the Default, just with 0.81\% increment. It means that Label Smooth is not competent for heavy noise and is unable to promote the model performance remarkably. The other comparative methods of combating noisy labels are highly dependent on the foreknowledge noise ratio. It can cherry-pick limited reliable and clean samples for training DNNs at the heavy noise rate, resulting in weak generalization in testing data. In addition, \textcolor{blue}{Fig. \ref{fig:asymmetric30}} illustrate the confusion matrices comparing test accuracy for Colon Pathology images classification with the Default, Co-Correcting, and our method, at the asymmetric noise rate of 30\%, respectively. From \textcolor{blue}{Fig. \ref{fig:asymmetric30}}, we can observe that our method mainly possesses the maximum values in each diagonal entry, which means it significantly surpasses the other methods and generally benefits each category to derive the more accurate predicted label. Although the noise rate was relatively high (30\%), our proposed technique still can promote the generalization performance of classification model on each category evenly.

\begin{figure*}[!t]
	\centering
	\includegraphics[width=1.8\columnwidth]{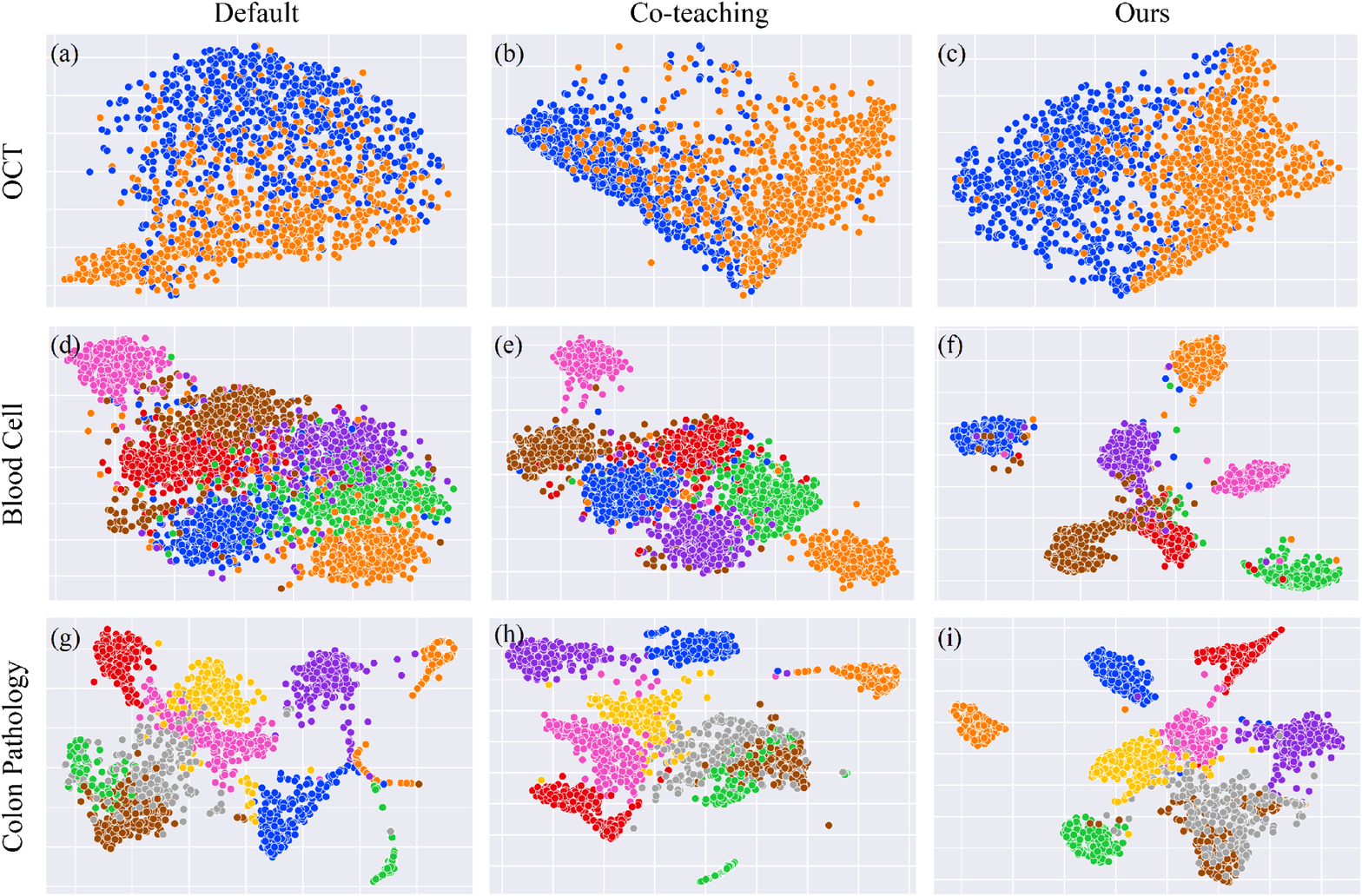}
	\vspace{0.0em}
	\caption{T-SNE visualizations of the feature vector after the last convolutional layer for each method (i.e., Default, Co-teaching, and Ours) under 30\% symmetric noise based on Retina OCT, Blood Cell, and Colon Pathology datasets, respectively.}
	\label{fig:tsne}
	\vspace{-1.5em}
\end{figure*}

\textbf{Comparison of the t-SNE visualization.} We furthermore implemented the t-SNE visualization for three representative methods, including the Default, Co-teaching and our method, based on Retina OCT, Blood Cell, and Colon Pathology datasets, respectively, which is shown in \textcolor{blue}{Fig. \ref{fig:tsne}}. Concretely, we utilized the best model in the last three epochs to perform the testing on the three medical datasets. During the testing process, feature vectors of test images were extracted from the last convolution layer of the well-trained model and were visualized by the t-SNE means. The visualization results manifest that the feature space of our method achieves the better clustering effect than that of the other two methods, especially on the Blood Cell dataset. We analyzed that our method contributed to an optimized embedding space with high-cohesion and low-coupling features, which can reduce the impact of label noise and boost the 
classification performance.

\subsection{Experiments on instance-dependent label noise}
\begin{figure*}[!t]
	\centering
	\includegraphics[width=1.8\columnwidth]{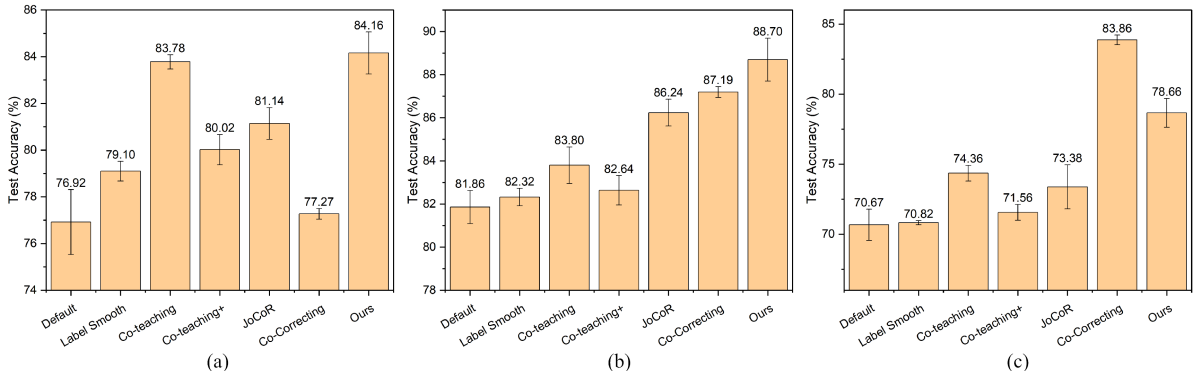}
	\vspace{-0.8em}
	\caption{Test accuracy of the comparative methods of combating noisy labels based on the instance-dependent noise. (a), (b), and (c) are for Retina OCT, Blood Cell, and Colon Pathology datasets, respectively.}
	\label{fig:IDN}
	\vspace{-1.5em}
\end{figure*}

This section presents the experimental results of aforementioned medical datasets on instance-dependent label noise. We construct instance-dependent label noise at ratio of around 20\% and \textcolor{blue}{Fig. \ref{fig:IDN}} respectively displays test accuracy of each anti-noise method on Retina OCT, Blood Cell, and Colon Pathology datasets. The corruption probability of instance-dependent label noise is assumed to be dependent on both the data features and class labels. It is more realistic noise model, but increasing the classification difficulty. As can be seen in \textcolor{blue}{Fig. \ref{fig:IDN}}, the performance of each anti-noise method is inferior when comparing results at the similar ratio of the instance-independent noise, correspondingly. However, all utilized methods can reduce the influence of instance-dependent label noise in various degrees. Our method achieves the top performance with 84.16\% and 88.70\% on Retina OCT and Blood Cell datasets, respectively. On Colon Pathology dataset, our method obtains the second-best performance with 78.66\% classification accuracy. It still surpasses the other methods with considerable increment other than Co-Correcting method. However, Co-Correcting proceeds co-train paradigm with dual-network, which costs nearly double computing resources and training time. In addition, Co-Correcting relies on the predetermined prior noise rate, which introduces computational complexity and estimation error. In a nutshell, our approach is competent across different noise patterns and is superior to the
comparison method in most cases. 

\subsection{Experimental results on real-world noisy dataset}
In addition to manually induced noise, we further evaluated generalization and robustness of our method on real-world noisy dataset, ANIMAL10N. The training details can be found in the implementation details section. To make the fair comparison, the number of epochs was changed to 150 (same as compared methods). Notably, the noise rate was set as 8\% for sample-selection style methods. Meanwhile, we report some performance of other popular anti-noise methods based on the ANIMAL10N given in original papers \cite{song2019selfie,gao2022bayesian}. We conducted our method on different backbones, including ResNet-18 and VGG-19. The test performance are demonstrated on \textcolor{blue}{Table \ref{tab: animal}}. We can conclude that our method is flexible to use without prior knowledge and independent of backbones.  It is superior to comparative methods both on ResNet-18 and VGG-19, with 82.79\% and 83.41\%, respectively.

\begin{table}[t!]
	\caption{Test accuracy of different anti-noise methods using the original paper settings on the ANIMAL10N dataset.}
	\begin{center}
		\label{tab: animal}
		\vspace{-0.8em}
		\resizebox{0.58 \columnwidth}{23mm}{
			\begin{tabular}{ccc} 
				\hline
				\hline
				Methods & Backbone & Accuracy \\ 
				\hline		
				\multirow{2}{*}{Default} & ResNet-18 & 75.09 \\
				~ & VGG-19 & 79.40 \\
				Label Smooth & ResNet-18 & 75.52 \\
				\multirow{2}{*}{Co-teaching} & ResNet-18 & 77.19 \\
				~ & VGG-19 & 80.20 \\
				Co-teaching$+$ & ResNet-18 & 76.50 \\
				JoCoR & ResNet-18 & 75.76 \\
				Co-Correcting & ResNet-18 & 75.91 \\
				ActiveBias & VGG-19 & 80.50 \\
				SELFIE & VGG-19 & 81.80 \\
				BLRM & VGG-19 & 82.60 \\
				\hline
				\multirow{2}{*}{Ours} & ResNet-18 & 82.79 \\
				~ & VGG-19 & \textbf{83.41} \\
				\hline
				\hline
			\end{tabular}
		}
	\end{center}
	\vspace{-2.3em}
\end{table}

\begin{table}[t!]
	\caption{The ablation study of each loss component based on the OCT dataset. The best results are highlighted.}
	\begin{center}
		\label{tab: ablation}
		\vspace{-0.8em}
		\resizebox{0.7 \columnwidth}{13mm}{
			\begin{tabular}{ccccccc} 
				\hline
				\hline
				\multirow{2}{*}{$\mathcal{L}_s$} & \multirow{2}{*}{$\mathcal{L}_m$} & \multirow{2}{*}{$\mathcal{L}_c$} & \multicolumn{4}{c}{Noise Rate
				} \\
				& & & 10\% & 20\% & 30\% & 40\%\\ 
				\hline
				\checkmark & $\times$ & $\times$ & 85.94 & 80.26 & 74.31 & 56.03 \\
				$\times$ & \checkmark & $\times$ & 85.67 & 84.73 & 80.31 & 58.27 \\
				\checkmark & \checkmark & $\times$ & 90.56 & 88.33 & 84.13 & 74.35 \\
				\checkmark & $\times$ & \checkmark & 92.80 & 89.09 & 88.07 & 82.69 \\
				$\times$ & \checkmark & \checkmark & 90.13 & 79.53 & 81.26 & 78.33 \\ 
				\checkmark & \checkmark & \checkmark & \textbf{94.02} & \textbf{91.24} & \textbf{89.56} & \textbf{85.18} \\ 						 			
				\hline
				\hline
			\end{tabular}
		}
	\end{center}
	\vspace{-2.0em}
\end{table}

\section{Discussion}
\subsection{Ablation study} 
\textcolor{blue}{Table \ref{tab: ablation}} shows ablation studies of each loss component, including supervised loss ($\mathcal{L}_s$), mixup loss ($\mathcal{L}_m$), and contrastive learning loss ($\mathcal{L}_c$), with different noise ratios.
The branch of contrastive learning loss, as a self-supervised learning technique, is for enhancing visual representation of DNNs and cannot independently output the predicted labels. As \textcolor{blue}{Table \ref{tab: ablation}} shows, compared with the baseline model which simply leverages $\mathcal{L}_s$, $\mathcal{L}_m$ gets better test accuracy through attention mixup operation under most noise ratios. 
Additionally, after integrating $\mathcal{L}_m$ into $\mathcal{L}_s$, the model further obtains significant improvements than any individual loss function at the varying noise rates. It proves that attention mixup technique contributes in boosting the generalization performance of model. 
Moreover, DNNs model guided by $\mathcal{L}_s$ along with $\mathcal{L}_c$, can also achieve better test accuracy under various noise levels. The more obvious gains indicate that contrastive learning loss is more effective at promoting the performance of model. Both attention mixup loss and contrastive learning loss can improve the classification accuracy individually.
Afterwards we combined $\mathcal{L}_m$ and $\mathcal{L}_c$ to constrain the model training together and the results can confirm the necessity and the importance of supervised loss ($\mathcal{L}_s$).  
In the end, the combination of all the three losses is validated optimal for this medical classification task, acquiring an accuracy of 94.02\%, 91.24\%, 89.56\%, and 85.18\% under the noise rate raising from 10\% to 40\%, respectively. To sum up, the results of experiments demonstrate the effectiveness and necessity of each loss component and the combination of them can achieve the optimal performance.

\subsection{Intra-class mixup vs. inter-class mixup}
\begin{figure*}[!t]
	\centering
	\includegraphics[width=1.89\columnwidth]{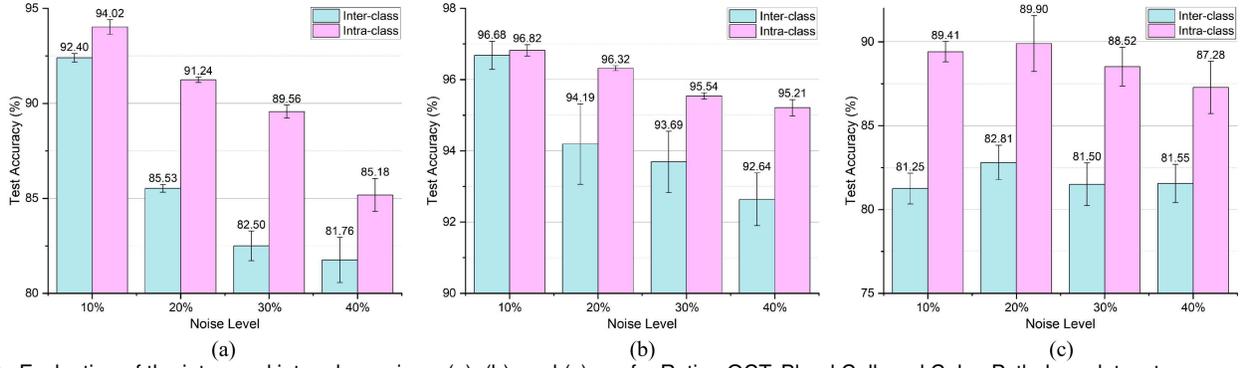}
	\vspace{-1.05em}
	\caption{Evaluation of the intra- and inter-class mixup. (a), (b), and (c) are for Retina OCT, Blood Cell, and Colon Pathology datasets, respectively.}
	\label{fig:mixup}
	\vspace{-1.3em}
\end{figure*}

Different from natural images, contents of medical images are more fine-grained. The similarity differences between inter-class and intra-class for medical images are mostly not obvious, which increases the difficulty of medical image classification tasks. Medical images with different categories or grades generally can be identified through the number (size and position) of lesions or the degree of morphological changes. Therefore, output images from inter-class mixup operation may results in some confusions between the distinguishable lesions or morphological changes and belonging labels. Intra-class mixup in the medical images may be an effective data augmentation manner, expanding the mapping range and boosting the model generalization. On the contrary, inter-class mixup operation may cause inconsistency between the image contents and given labels, bringing in adverse effects on model generalization. We conducted evaluation of the intra- and inter-class mixup operations respectively based on the three above-mentioned medical datasets. \textcolor{blue}{Fig. \ref{fig:mixup}} illustrates the histograms of test accuracy for comparison experiments. All the experimental results can support that intra-class mixup operation is more beneficial than the others.
Take Retina OCT dataset for example, performance with inter-class mixup overpasses it with intra-class mixup, by 1.62\%, 5.71\%, 7.06\%, and 3.42\% gains, under the noise rate raising from 10\% to 40\%, respectively. 

\subsection{The influence of mixup size}
\begin{figure}[!t]
	\centering
	\includegraphics[width=1.0\columnwidth]{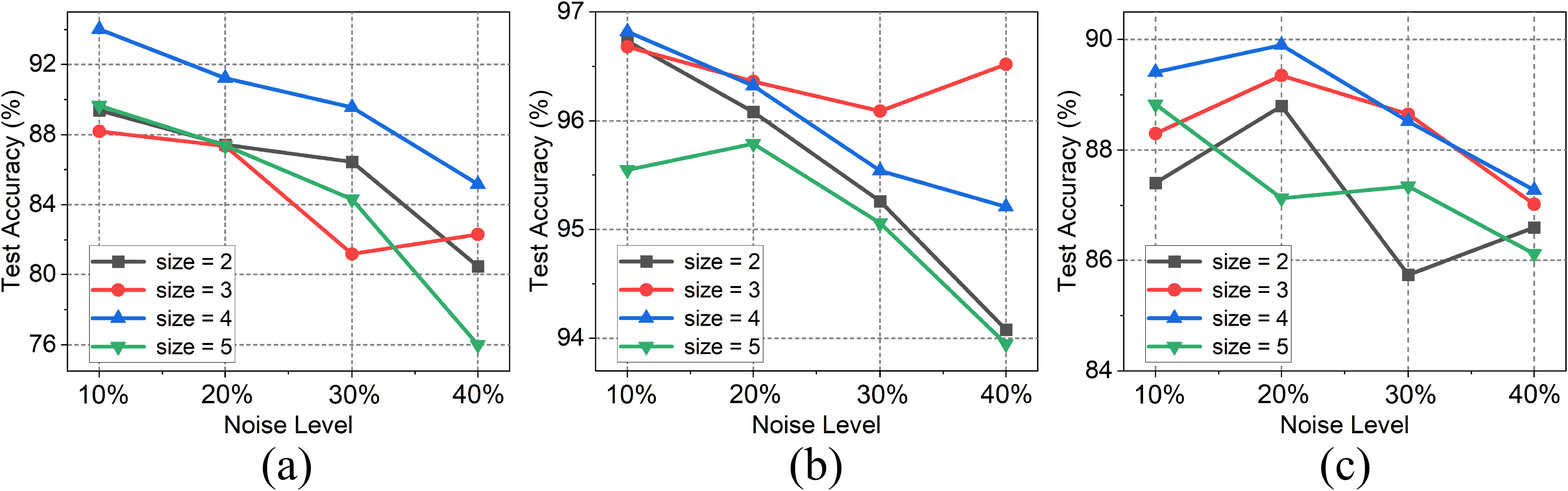}
	\vspace{-1.8em}
	\caption{The influence of mixup size. (a), (b), and (c) are for Retina OCT, Blood Cell, and Colon Pathology datasets, respectively.}
	\label{fig:mixup_size}
	\vspace{-1.3em}
\end{figure}

We conducted hyper-parameter sensitivity analysis experiments to determine the optimal mixup size value for each medical dataset. Note that, when investigating different sizes of mixup operation, we need to keep the batch size stable to eliminate influence of varying batch size. As the batch size being equal to the product of mixup size and the number of sampled categories in the batch, when we changed the mixup size, we adjusted the corresponding batch size slightly. The experimental results based on the three noisy medical datasets under different noise ratios (with mixup size 2, 3, 4, and 5) are illustrated in \textcolor{blue}{Fig. \ref{fig:mixup_size}}. We can observe that the best test accuracy cannot be achieved with both small and large mixup sizes. The attention noisy features mixup with small mixup size is hard to generate clean feature representation, which is limited in mitigating the impact of label noise. The attention noisy features mixup with large mixup size corresponds to hallucination representation which is probably completely deviated from the real-world feature representation. The performance generally involves a comprise between the clean feature representation and pure hallucination representation. In \textcolor{blue}{Fig. \ref{fig:mixup_size} a}, for Retina OCT dataset, the mixup size of 4 achieves the highest test accuracies at different noise ratios.
Regarding Blood Cell (\textcolor{blue}{Fig. \ref{fig:mixup_size} b}) and Colon Pathology datasets (\textcolor{blue}{Fig. \ref{fig:mixup_size} c}), although there is no clear winner for the specific mixup size, the size of 4 is still competent and gets comparable test accuracies.
Therefore, we set the mixup size to 4 for all of medical datasets.

\subsection{The architecture of noisy weight}
We conducted comparison experiments followed the architecture of FC($N$)-ReLU-Sig ($N$ is the number of FC layer, $N\in\left[1,2,3\right]$), to seek for appropriate architecture for the noisy weight module. \textcolor{blue}{Table \ref{tab: noisy_weight}} demonstrates the multiple architectures, dimensions (dim.), and the test accuracy on the three medical datasets. From \textcolor{blue}{Table \ref{tab: noisy_weight}}, regarding the Retina OCT dataset, we empirically find that in our cases fewer or more FC layers in the noisy weight cannot achieve the best performance under each noise level. In contrast, the noisy weight in the architecture of FC-FC-ReLU-Sig is more effective for classification than others. Regarding to Blood Cell dataset, noisy weight with FC(3)-ReLU-Sig architecture obtains better performance under each noise ratio by a small margin, compared with FC-FC-ReLU-Sig. Nevertheless, for Colon Pathology dataset, there is no any architecture of noisy weight occupying the dominant position at the varying noise ratios. In summary, we utilized FC(2)-ReLU-Sig as the noisy weight for both Retina OCT and Colon Pathology datasets, FC(3)-ReLU-Sig for Blood Cell dataset, respectively.

\begin{table}[t!]
	\caption{The test accuracy on three datasets, for choosing the best noisy weight of mixup operation.}
	\begin{center}
		\label{tab: noisy_weight}
		\vspace{-2.5em}
		\resizebox{1.0 \columnwidth}{16mm}{
			\begin{tabular}{ccccccc} 
				\hline
				\hline
				\multirow{2}{*}{Datasets} & \multirow{2}{*}{Architecture} & \multirow{2}{*}{Dim.} & \multicolumn{4}{c}{Noise Rate
				} \\
				& & & 10\% & 20\% & 30\% & 40\%\\ 
				\hline
				\multirow{3}{*}{Retina OCT} & FC(1)-ReLU-Sig & 512*4,4 & 91.49 & 88.60 & 88.09 & 79.11 \\
				~ & FC(2)-ReLU-Sig & 512*4,512,4 & \textbf{94.02} & \textbf{91.24} & \textbf{89.56} & \textbf{85.18} \\
				~ & FC(3)-ReLU-Sig & 512*4,512,512,4 & 91.42 & 85.04 & 85.44 & 81.02 \\
				\hline
				\multirow{3}{*}{Blood Cell} & FC(1)-ReLU-Sig & 512*4,4 & 96.00 & 94.39 & 94.67 & 95.44 \\
				~ & FC(2)-ReLU-Sig & 512*4,512,4 & 96.82 & 96.32 & 95.54 & 95.21 \\
				~ & FC(3)-ReLU-Sig & 512*4,512,512,4 & \textbf{96.89} & \textbf{97.01} & \textbf{96.17} & \textbf{95.73} \\
				\hline
				\multirow{3}{*}{Colon Pathology} & FC(1)-ReLU-Sig & 512*4,4 & \textbf{89.62} & 85.35 & 84.28 & 85.53 \\
				~ & FC(2)-ReLU-Sig & 512*4,512,4 & 89.41 & \textbf{89.90} & 88.52 & \textbf{87.28} \\
				~ & FC(3)-ReLU-Sig & 512*4,512,512,4 & 89.10 & 88.05 & \textbf{88.60} & 87.28 \\						 			
				\hline
				\hline
			\end{tabular}
		}
	\end{center}
	\vspace{-1.0em}
\end{table}

\begin{figure}[!t]
	\centering
	\includegraphics[width=1.0\columnwidth]{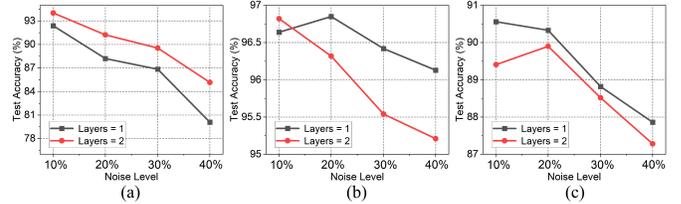}
	\vspace{-1.5em}
	\caption{The influence of projection layers. (a), (b), and (c) are for Retina OCT, Blood Cell, and Colon Pathology datasets, respectively.}
	\label{fig:projection}
	\vspace{-1.5em}
\end{figure}

\subsection{The analysis of projection head}
We further investigated the influence of different layers for projection head. Concretely, experimental results with projection head consisting of one-layer or two-layer neural networks, based on the three medical datasets, were illustrated in \textcolor{blue}{Fig. \ref{fig:projection}}. In our training procedure, embedding features from feature extractor were further fed into the projection head which is indeed complementary deep convolution. On the Retina OCT dataset, classification accuracy from the projection head with two layers has the best performance for all of noise levels. We speculate that in this binary classification task, the positive cover three categories of diseases (choroidal neovascularization, diabetic macular edema, and drusen), which expands the data diversity and needs deeper structures to explore feature representation. By contrast, projection head with one layer is mainly advantageous on the remaining two medical datasets. Although the two datasets possess multiple categories and more training samples, these categories have more discriminative power which can be supported by their classification performance. To sum up, we utilized one-layer projection head for the Retina OCT dataset while two-layer projection head for the other two medical datasets in our experiments.
 
\section{Conclusion}
This paper proposes an effective method for mitigating the adverse effects of label noise using self-supervised and attention mixup techniques. Self-supervised learning can boost visual representations by leveraging the whole training datasets, especially the noisy data. Moreover, noisy attention mixup can reduce the influence of noisy data, enlarging the mapping range. Extensive experiments of three synthetic noisy medical datasets demonstrate the superiority of our method. Concurrently, we verified the feasibility of our method on the real-world noisy datasets and further confirmed our method’s stability and robustness.
Notably, our approach do not need any prior, such as noise rate and clean validation set. It opens up the possibility of dealing with noisy labels in the real-world medical classification applications.

\bibliographystyle{IEEEtran}
\bibliography{refs}

\begin{thebibliography}{10}
\providecommand{\url}[1]{#1}
\csname url@samestyle\endcsname
\providecommand{\newblock}{\relax}
\providecommand{\bibinfo}[2]{#2}
\providecommand{\BIBentrySTDinterwordspacing}{\spaceskip=0pt\relax}
\providecommand{\BIBentryALTinterwordstretchfactor}{4}
\providecommand{\BIBentryALTinterwordspacing}{\spaceskip=\fontdimen2\font plus
\BIBentryALTinterwordstretchfactor\fontdimen3\font minus
  \fontdimen4\font\relax}
\providecommand{\BIBforeignlanguage}[2]{{%
\expandafter\ifx\csname l@#1\endcsname\relax
\typeout{** WARNING: IEEEtran.bst: No hyphenation pattern has been}%
\typeout{** loaded for the language `#1'. Using the pattern for}%
\typeout{** the default language instead.}%
\else
\language=\csname l@#1\endcsname
\fi
#2}}
\providecommand{\BIBdecl}{\relax}
\BIBdecl

\bibitem{karimi2020deep}
D.~Karimi, H.~Dou, S.~K. Warfield, and A.~Gholipour, ``Deep learning with noisy
  labels: Exploring techniques and remedies in medical image analysis,''
  \emph{Medical image analysis}, vol.~65, p. 101759, 2020.

\bibitem{cheplygina2018crowd}
V.~Cheplygina and J.~P. Pluim, ``Crowd disagreement about medical images is
  informative,'' in \emph{Intravascular imaging and computer assisted stenting
  and large-scale annotation of biomedical data and expert label
  synthesis}.\hskip 1em plus 0.5em minus 0.4em\relax Springer, 2018, pp.
  105--111.

\bibitem{wang2018tienet}
X.~Wang, Y.~Peng, L.~Lu, Z.~Lu, and R.~M. Summers, ``Tienet: Text-image
  embedding network for common thorax disease classification and reporting in
  chest x-rays,'' in \emph{Proceedings of the IEEE conference on computer
  vision and pattern recognition}, 2018, pp. 9049--9058.

\bibitem{wang2020seminll}
Z.~Wang, J.~Jiang, B.~Han, L.~Feng, B.~An, G.~Niu, and G.~Long, ``Seminll: A
  framework of noisy-label learning by semi-supervised learning,'' \emph{arXiv
  preprint arXiv:2012.00925}, 2020.

\bibitem{arazo2020pseudo}
E.~Arazo, D.~Ortego, P.~Albert, N.~E. O’Connor, and K.~McGuinness,
  ``Pseudo-labeling and confirmation bias in deep semi-supervised learning,''
  in \emph{2020 International Joint Conference on Neural Networks
  (IJCNN)}.\hskip 1em plus 0.5em minus 0.4em\relax IEEE, 2020, pp. 1--8.

\bibitem{kuznetsova2020open}
A.~Kuznetsova, H.~Rom, N.~Alldrin, J.~Uijlings, I.~Krasin, J.~Pont-Tuset,
  S.~Kamali, S.~Popov, M.~Malloci, A.~Kolesnikov \emph{et~al.}, ``The open
  images dataset v4,'' \emph{International Journal of Computer Vision}, vol.
  128, no.~7, pp. 1956--1981, 2020.

\bibitem{dgani2018training}
Y.~Dgani, H.~Greenspan, and J.~Goldberger, ``Training a neural network based on
  unreliable human annotation of medical images,'' in \emph{2018 IEEE 15th
  International Symposium on Biomedical Imaging (ISBI 2018)}.\hskip 1em plus
  0.5em minus 0.4em\relax IEEE, 2018, pp. 39--42.

\bibitem{park2021provable}
S.~Park, J.~Lee, C.~Yun, and J.~Shin, ``Provable memorization via deep neural
  networks using sub-linear parameters,'' in \emph{Conference on Learning
  Theory}.\hskip 1em plus 0.5em minus 0.4em\relax PMLR, 2021, pp. 3627--3661.

\bibitem{han2018co}
B.~Han, Q.~Yao, X.~Yu, G.~Niu, M.~Xu, W.~Hu, I.~Tsang, and M.~Sugiyama,
  ``Co-teaching: Robust training of deep neural networks with extremely noisy
  labels,'' \emph{Advances in neural information processing systems}, vol.~31,
  2018.

\bibitem{liu2015classification}
T.~Liu and D.~Tao, ``Classification with noisy labels by importance
  reweighting,'' \emph{IEEE Transactions on pattern analysis and machine
  intelligence}, vol.~38, no.~3, pp. 447--461, 2015.

\bibitem{goldberger2016training}
J.~Goldberger and E.~Ben-Reuven, ``Training deep neural-networks using a noise
  adaptation layer,'' 2016.

\bibitem{jenni2018deep}
S.~Jenni and P.~Favaro, ``Deep bilevel learning,'' in \emph{Proceedings of the
  European conference on computer vision (ECCV)}, 2018, pp. 618--633.

\bibitem{song2019selfie}
H.~Song, M.~Kim, and J.-G. Lee, ``Selfie: Refurbishing unclean samples for
  robust deep learning,'' in \emph{International Conference on Machine
  Learning}.\hskip 1em plus 0.5em minus 0.4em\relax PMLR, 2019, pp. 5907--5915.

\bibitem{he2020momentum}
K.~He, H.~Fan, Y.~Wu, S.~Xie, and R.~Girshick, ``Momentum contrast for
  unsupervised visual representation learning,'' in \emph{Proceedings of the
  IEEE/CVF conference on computer vision and pattern recognition}, 2020, pp.
  9729--9738.

\bibitem{chen2020simple}
T.~Chen, S.~Kornblith, M.~Norouzi, and G.~Hinton, ``A simple framework for
  contrastive learning of visual representations,'' in \emph{International
  conference on machine learning}.\hskip 1em plus 0.5em minus 0.4em\relax PMLR,
  2020, pp. 1597--1607.

\bibitem{grill2020bootstrap}
J.-B. Grill, F.~Strub, F.~Altch{\'e}, C.~Tallec, P.~Richemond, E.~Buchatskaya,
  C.~Doersch, B.~Avila~Pires, Z.~Guo, M.~Gheshlaghi~Azar \emph{et~al.},
  ``Bootstrap your own latent-a new approach to self-supervised learning,''
  \emph{Advances in neural information processing systems}, vol.~33, pp.
  21\,271--21\,284, 2020.

\bibitem{peng2020suppressing}
X.~Peng, K.~Wang, Z.~Zeng, Q.~Li, J.~Yang, and Y.~Qiao, ``Suppressing
  mislabeled data via grouping and self-attention,'' in \emph{European
  Conference on Computer Vision}.\hskip 1em plus 0.5em minus 0.4em\relax
  Springer, 2020, pp. 786--802.

\bibitem{jing2020self}
L.~Jing and Y.~Tian, ``Self-supervised visual feature learning with deep neural
  networks: A survey,'' \emph{IEEE transactions on pattern analysis and machine
  intelligence}, vol.~43, no.~11, pp. 4037--4058, 2020.

\bibitem{oord2018representation}
A.~v.~d. Oord, Y.~Li, and O.~Vinyals, ``Representation learning with
  contrastive predictive coding,'' \emph{arXiv preprint arXiv:1807.03748},
  2018.

\bibitem{kermany2018identifying}
D.~S. Kermany, M.~Goldbaum, W.~Cai, C.~C. Valentim, H.~Liang, S.~L. Baxter,
  A.~McKeown, G.~Yang, X.~Wu, F.~Yan \emph{et~al.}, ``Identifying medical
  diagnoses and treatable diseases by image-based deep learning,'' \emph{Cell},
  vol. 172, no.~5, pp. 1122--1131, 2018.

\bibitem{acevedo2020dataset}
A.~Acevedo, A.~Merino, S.~Alf{\'e}rez, {\'A}.~Molina, L.~Bold{\'u}, and
  J.~Rodellar, ``A dataset of microscopic peripheral blood cell images for
  development of automatic recognition systems,'' \emph{Data in brief},
  vol.~30, 2020.

\bibitem{kather2019predicting}
J.~N. Kather, J.~Krisam, P.~Charoentong, T.~Luedde, E.~Herpel, C.-A. Weis,
  T.~Gaiser, A.~Marx, N.~A. Valous, D.~Ferber \emph{et~al.}, ``Predicting
  survival from colorectal cancer histology slides using deep learning: A
  retrospective multicenter study,'' \emph{PLoS medicine}, vol.~16, no.~1, p.
  e1002730, 2019.

\bibitem{song2022learning}
H.~Song, M.~Kim, D.~Park, Y.~Shin, and J.-G. Lee, ``Learning from noisy labels
  with deep neural networks: A survey,'' \emph{IEEE Transactions on Neural
  Networks and Learning Systems}, 2022.

\bibitem{liu2021co}
J.~Liu, R.~Li, and C.~Sun, ``Co-correcting: noise-tolerant medical image
  classification via mutual label correction,'' \emph{IEEE Transactions on
  Medical Imaging}, vol.~40, no.~12, pp. 3580--3592, 2021.

\bibitem{szegedy2016rethinking}
C.~Szegedy, V.~Vanhoucke, S.~Ioffe, J.~Shlens, and Z.~Wojna, ``Rethinking the
  inception architecture for computer vision,'' in \emph{Proceedings of the
  IEEE conference on computer vision and pattern recognition}, 2016, pp.
  2818--2826.

\bibitem{yu2019does}
X.~Yu, B.~Han, J.~Yao, G.~Niu, I.~Tsang, and M.~Sugiyama, ``How does
  disagreement help generalization against label corruption?'' in
  \emph{International Conference on Machine Learning}.\hskip 1em plus 0.5em
  minus 0.4em\relax PMLR, 2019, pp. 7164--7173.

\bibitem{wei2020combating}
H.~Wei, L.~Feng, X.~Chen, and B.~An, ``Combating noisy labels by agreement: A
  joint training method with co-regularization,'' in \emph{Proceedings of the
  IEEE/CVF Conference on Computer Vision and Pattern Recognition}, 2020, pp.
  13\,726--13\,735.

\bibitem{he2016deep}
K.~He, X.~Zhang, S.~Ren, and J.~Sun, ``Deep residual learning for image
  recognition,'' in \emph{Proceedings of the IEEE conference on computer vision
  and pattern recognition}, 2016, pp. 770--778.

\bibitem{gao2022bayesian}
M.~Gao, X.~Feng, M.~Geng, Z.~Jiang, L.~Zhu, X.~Meng, C.~Zhou, Q.~Ren, and
  Y.~Lu, ``Bayesian statistics-guided label refurbishment mechanism: Mitigating
  label noise in medical image classification,'' \emph{Medical Physics},
  vol.~49, no.~9, pp. 5899--5913, 2022.

\end{thebibliography}

\end{document}